\documentclass[runningheads]{llncs}
\usepackage[T1]{fontenc}
\usepackage{graphicx}
\usepackage{amsmath}
\usepackage{amssymb}
\usepackage{tikz}
\usepackage{xspace}
\usepackage{algorithm}
\usepackage{algorithmic}
\usepackage{bbding}

\begin{document}
\title{Improving Diffusion Models for the Traveling Salesman Problem (TSP) by Leveraging the Structure of the Solution Space}
\titlerunning{Improving Diffusion Models for the TSP}
%
\author{Micka\"el Basson\inst{1,2,3,4}\Envelope \and
Philippe Preux\inst{1,2,3, 4}} 
%
\authorrunning{M. Basson, P. Preux}
%
\institute{Université de Lille, France \and 
CNRS, France \and
Inria, France \and 
UMR 9198-CRIStAL, Lille, France \\
\email{\{mickael.basson, philippe.preux\}@inria.fr}}
%

\newcommand{\XYZ}{IDEQ\xspace}
\newcommand{\XYZfull}{IDEQ (constrained Inverse Diffusion and EQuivalence class-based training of diffusion models for combinatorial optimization)\xspace}

\maketitle              
\begin{abstract}
In this paper we show how leveraging the structure of the solution space of the Traveling Salesman Problem (TSP) can lead to a dramatic improvement of the performance of state of the art diffusion-based neural solvers. Building on recent approaches of DIFUSCO and T2TCO which pipeline a diffusion-based solution generation with a local search procedure, we propose  \XYZfull. \XYZ improves the quality of the solutions by leveraging the constrained structure of the TSP state space. Indeed, the solution space consists of locally optimal Hamiltonian tours which is a much smaller space than the space of adjacency matrices used in previous works. Also, the local search procedure defines an equivalence class of Hamiltonian tours: all elements of this equivalence class reach the same local optimum after the application of the local search. This should be aligned with the supervised training objective of the diffusion. \XYZ addresses these two points. Our experiments show that \XYZ achieves 0.3\% to 0.4\% optimality gap on TSP instances made of 500 cities, and 0.5\% to 0.6\% optimality gap on TSP instances with 1000 cities. This sets a new SOTA for neural based methods solving the TSP. \XYZ also performs well on the instances of the TSPlib, a reference benchmark in the TSP community, outside of the training distribution, with optimality gaps ranging from 0.9 to 1.1 \%.
\end{abstract}
\section{Introduction}

Recent years have seen a surge in machine learning models to solve combinatorial optimization (CO) problems. The field of combinatorial optimization is a historical field of research and application in computer science. After decades of progress, very efficient algorithms exist to provide exact solutions or approximate solutions to many CO problems. The Traveling Salesman Problem (TSP) stands as a prominent example of this fact: the TSP is very appealing as it is very simple to understand, it has a wide range of applications, and we are able to solve exactly rather large instances of this problem on a mere laptop (an instance defined over a few thousands cities can be solved within one hour). We also have approximate algorithms that are able to find tours that are very close to optimality: LKH3 \cite{lkh3} can solve instances of 40,000 cities in about one hour on a laptop, however we can not have any guarantee about the optimality of the resulting tour. 

Because of its appeal, the TSP has also drawn the attention of researchers in deep neural networks in the recent years. If the first attempts had difficulties solving even small TSP instances of a few dozen cities, progress has been made (see e.g. references in \cite{KoolAM}). In this paper we build on these previous works and go a step further. In terms of experimental results, we provide the new state-of-the-art neural results on TSP instances of size up to 7397 cities. 
These results are competitive with the best heuristic approaches for the TSP in terms of the quality of the solution, and competitive with exact solvers in terms of computation time. 

Our approach, \XYZ, relies on diffusion-based models
, and particularly the recent DIFUSCO and T2TCO \cite{difusco,T2TCO}. Solving a supervised learning problem, \XYZ aims at learning an optimal tour for a given TSP instance. 
With regards to DIFUSCO and T2TCO, we improve the inference 
by leveraging the highly constrained structure of the solution space of the TSP.
We also replace the later stage of the training curriculum of DIFUSCO by adding a simplified objective replacing the unique instance-conditioned ground truth solution by a uniform distribution over a carefully chosen equivalence class of the ground truth solution.

The paper is organized as follows: after a review of the related literature, we describe \XYZ, a new diffusion-based combinatorial optimization solver for the TSP. In applied CO, 
the criterion is the experimental performance. So we took a great care to investigate the experimental performance of \XYZ. 
Following the CO community practices, we experimented \XYZ on the TSPlib which is a well-known benchmark on which we report state of the art performance using a neural net based approach. 
Then, we perform a detailed analysis of the components of \XYZ to disentangle the contribution of each.


\section{Background and Related Works}

\subsection{About the TSP}

Let us consider the basic and usual definition of the Traveling Salesman Problem (see e.g. \cite{johnsonBiblicalArticle}). An instance of the TSP is defined in the Euclidean plane by a set of $N$ cities. $N$ is also known as the ``size'' of the instance. We will denote TSP-$N$ a TSP instance of size $N$. An instance may be defined either by the location of the cities or by a distance matrix between the cities. The goal is to find an ordering of the cities such that the associated tour has a minimal length. There exists many variations of this definition in Euclidean and non Euclidean spaces. In this paper, TSP refers to this particular family of 
Euclidean instances. The TSP is a well-known example of a NP-hard problem: the time/memory requirements to solve an instance of size $N$ grow in $O (\exp{(N)})$ so that the size of the instances that can be solved within a reasonable amount of time is very limited. For instance, today, the state-of-the-art exact solver Concorde \cite{concorde} solves an instance of the TSP of size $N=10^3$ in about 10 minutes on a laptop. For larger instances, one has to use heuristics that have no formal guarantee to find an optimal tour. 
The quality of a tour may be measured by its ``optimality gap'', defined by $\frac{\mbox{length of the tour} - \mbox{optimal tour length}}{\mbox{optimal tour length}}$: it is a non-negative real number equal to 0 only for an optimal tour\footnote{In the case of large instances we use the best solution provided by heuristics as a proxy for the optimal solution. In this case we may encounter situations where the optimality gap is negative. Indeed, if the heuristics gives a non-optimal solution any solution between optimality and the heuristics one has a negative optimality gap. This has very limited impact in practice.}. 
There is no way to assess a priori the difficulty of an instance: we know for sure that its size says nothing about it: for any $N$, it is obvious to design an instance which is trivial to solve.

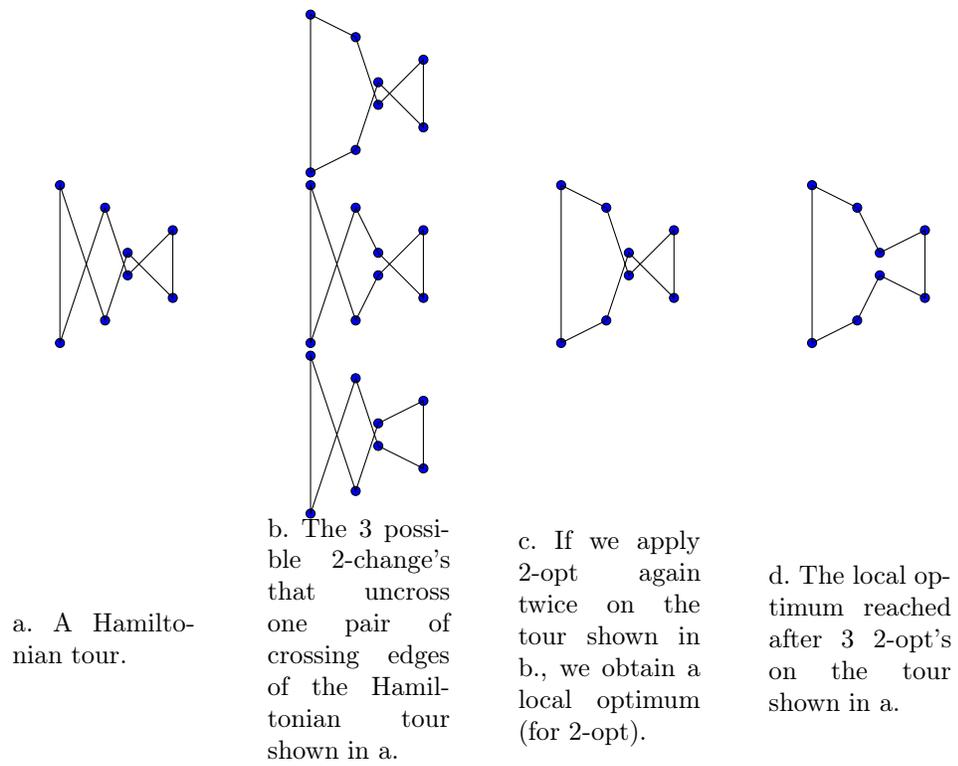
\begin{figure*}
  \begin{tabular}{ccc}
    \setlength{\tabcolsep}{0pt}
    \begin{minipage}{.3\textwidth}
      \begin{tikzpicture}[scale=2]
        \filldraw[fill=blue] (0.1,0) circle (0.02);
        \filldraw[fill=blue] (0.1,0.7) circle (0.02);
        \filldraw[fill=blue] (0.3,0.1) circle (0.02);
        \filldraw[fill=blue] (0.3,0.6) circle (0.02);
        \filldraw[fill=blue] (0.4,0.3) circle (0.02);
        \filldraw[fill=blue] (0.4,0.4) circle (0.02);
        \filldraw[fill=blue] (0.6,0.2) circle (0.02);
        \filldraw[fill=blue] (0.6,0.5) circle (0.02);
        \draw (0.1,0) -- (0.1,0.7) -- (0.3,0.1) -- (0.4,0.4) -- (0.6,0.2) -- (0.6,0.5) -- (0.4,0.3) -- (0.3,0.6) -- (0.1,0);
      \end{tikzpicture}
    \end{minipage}
    &
    \begin{minipage}{.4\textwidth}
      \begin{tikzpicture}[scale=2]
        \filldraw[fill=blue] (0.1,0) circle (0.02);
        \filldraw[fill=blue] (0.1,0.7) circle (0.02);
        \filldraw[fill=blue] (0.3,0.1) circle (0.02);
        \filldraw[fill=blue] (0.3,0.6) circle (0.02);
        \filldraw[fill=blue] (0.4,0.3) circle (0.02);
        \filldraw[fill=blue] (0.4,0.4) circle (0.02);
        \filldraw[fill=blue] (0.6,0.2) circle (0.02);
        \filldraw[fill=blue] (0.6,0.5) circle (0.02);
        \draw (0.1,0) -- (0.3,0.1) -- (0.4,0.4) -- (0.6,0.2) -- (0.6,0.5) -- (0.4,0.3) -- (0.3,0.6) -- (0.1,0.7) -- (0.1,0);
      \end{tikzpicture} \hspace{.5cm}
      \begin{tikzpicture}[scale=2]
        \filldraw[fill=blue] (0.1,0) circle (0.02);
        \filldraw[fill=blue] (0.1,0.7) circle (0.02);
        \filldraw[fill=blue] (0.3,0.1) circle (0.02);
        \filldraw[fill=blue] (0.3,0.6) circle (0.02);
        \filldraw[fill=blue] (0.4,0.3) circle (0.02);
        \filldraw[fill=blue] (0.4,0.4) circle (0.02);
        \filldraw[fill=blue] (0.6,0.2) circle (0.02);
        \filldraw[fill=blue] (0.6,0.5) circle (0.02);
        \draw (0.1,0) -- (0.1,0.7) -- (0.3,0.1) -- (0.4,0.3) -- (0.6,0.5) -- (0.6,0.2) -- (0.4,0.4) -- (0.3,0.6) -- (0.1,0);
      \end{tikzpicture}\hspace{.5cm}
      \begin{tikzpicture}[scale=2]
        \filldraw[fill=blue] (0.1,0) circle (0.02);
        \filldraw[fill=blue] (0.1,0.7) circle (0.02);
        \filldraw[fill=blue] (0.3,0.1) circle (0.02);
        \filldraw[fill=blue] (0.3,0.6) circle (0.02);
        \filldraw[fill=blue] (0.4,0.3) circle (0.02);
        \filldraw[fill=blue] (0.4,0.4) circle (0.02);
        \filldraw[fill=blue] (0.6,0.2) circle (0.02);
        \filldraw[fill=blue] (0.6,0.5) circle (0.02);
        \draw (0.1,0) -- (0.1,0.7) -- (0.3,0.1) -- (0.4,0.4) -- (0.6,0.5) -- (0.6,0.2) -- (0.4,0.3) -- (0.3,0.6) -- (0.1,0);
      \end{tikzpicture}
      \vspace{.5cm}
    \end{minipage}
    &
    \begin{minipage}{.3\textwidth}
      \begin{tikzpicture}[scale=2]
        \filldraw[fill=blue] (0.1,0) circle (0.02);
        \filldraw[fill=blue] (0.1,0.7) circle (0.02);
        \filldraw[fill=blue] (0.3,0.1) circle (0.02);
        \filldraw[fill=blue] (0.3,0.6) circle (0.02);
        \filldraw[fill=blue] (0.4,0.3) circle (0.02);
        \filldraw[fill=blue] (0.4,0.4) circle (0.02);
        \filldraw[fill=blue] (0.6,0.2) circle (0.02);
        \filldraw[fill=blue] (0.6,0.5) circle (0.02);
        \draw (0.1,0) -- (0.3,0.1) -- (0.4,0.3) -- (0.6,0.2) -- (0.6,0.5) -- (0.4,0.4) -- (0.3,0.6) -- (0.1,0.7) -- (0.1,0);
      \end{tikzpicture}
    \end{minipage} \\
    \hspace*{-15mm}
    \begin{minipage}{.25\textwidth}
      (a).\@ A Hamiltonian tour.
    \end{minipage} &
    \hspace*{-15mm}
    \begin{minipage}{.4\textwidth}
      (b).\@ The 3 possible 2-change's that uncross one pair of crossing edges of the Hamiltonian tour shown in (a) (note: there are other possible 2-changes that do not uncross edges, they have not been displayed for simplicity)
    \end{minipage} &
    \hspace*{-15mm}
    \begin{minipage}{.25\textwidth}
      (c).\@ The local optimum reached after 2 2-opt's on the tour shown in (a).
    \end{minipage} \\ 
  \end{tabular}
  \caption{Illustration of 2-change and 2-opt operators on a Hamiltonian tour.}
  \label{fig:2opt}
\end{figure*}

An optimal tour goes once and only once per city: it is a Hamiltonian tour. Consequently, the search space may be reduced to the set of Hamiltonian tours. 
Given a Hamiltonian tour, there is a very simple heuristic that can decrease its length  known as ``2-opt'' (see Fig.\@ \ref{fig:2opt}). Given a Hamiltonian tour (Fig.\@ \ref{fig:2opt}.a), a ``2-change'' consists in removing a pair of edges of the tour and reconstructing another tour (a single tour may be reconstructed). In 2D, when the edges of the pair cross each other, applying a 2-change uncrosses the edges (Fig.\@ \ref{fig:2opt}.b): this is guaranteed to decrease the length of the tour. 2-opt considers all pairs of crossing edges and disentangles the pair that most decreases the length of the resulting tour (Fig.\@ \ref{fig:2opt}.c). 2-opt is both very simple and rather effective in optimizing a tour. In practice, it is customary to pipeline a first heuristic that produces a Hamiltonian tour, and then post-process it by iterating 2-opt to obtain a disentangled tour. There exists other such local optimizations, like the Lin-Kernighan operator (lk-opt) \cite{lk} and its more recent version LKH3 \cite{lkh3} that are more complex in terms of algorithm, though very efficient and effective in practice. For a (still) relevant survey, we refer the interested reader to \cite{johnsonBiblicalArticle}. In this paper, we will restrict ourselves to the use of the 2-opt since it is simple, fast, and yet efficient.

\subsection{Denoising Diffusion Models}

Denoising diffusion models (introduced in \cite{Sohl}, and then \cite{Ho,Song,Nichol,Dhariwal}) aim at being able to sample a complex data distribution. The idea is to start from an easy to sample distribution and then going through a set of iterative Markovian transformations which mimics the reverse of a diffusion process, to sample from more and more complex distributions up to reaching the target complex distribution. The training of these models is done in a supervised manner.
This family of generative models has now become SOTA for (conditioned) image and video generation \cite{stableDiff,imgvid1,imgvid2,imgvid3}. They have also achieved great performance in areas like audio synthesis \cite{audio1,audio2}, molecule generation \cite{mol1,Vignac,Hoogeboom2}, text generation \cite{Austin}, and combinatorial optimization \cite{difusco}.
While most of denoising diffusion models are defined on a continuous state space, discrete space models have also been developed (initially introduced by \cite{Sohl} and further expanded by \cite{Austin,Hoogeboom}).

\subsection{Diffusion Models for Combinatorial Optimization}

A solution to a TSP can be represented by an $N\times{}N$ adjacency matrix $A = (a_{i,j})$ where $a_{i,j}=1$ if there exists a link between city $i$ and city $j$, 0 otherwise. 
One may soften this definition and define $a_{i,j}$ as the probability that there exits a link between $i$ and $j$. This produces a heatmap which may be seen as a stochastic adjacency matrix.


To predict the adjacency matrix of graph-based CO problems, diffusion models have been introduced by DIFUSCO \cite{difusco}. In DIFUSCO, a denoising diffusion process is used to predict the heatmap of the solution for a given TSP instance starting from an easy to sample distribution (uniform distribution with probability $1/2$ for all off-diagonal terms and $0$ on the diagonal). The heatmap is converted into an adjacency matrix with a deterministic tour reconstruction operator.  
DIFUSCO can be used with either continuous, discrete, or discretized diffusion. For the TSP, the best results were obtained with a discrete diffusion. 

The training is done in a supervised manner on large synthetic datasets.
One such dataset is made of thousands up to a million random instances of a certain size $N$ labeled with their optimal (or best known) tour.
Training requires significant computational resources to create the dataset (we need to compute the optimal tour of each instance of the dataset) and then to train the network. DIFUSCO is trained by curriculum learning: the network is first trained on random TSP-100 2D instances. 
Then, this neural network is further trained on TSP-500 instances. Then again, the network is trained further on TSP-1000 instances, and so on. We call ``TSP-$N$ checkpoint'' the weights of the neural network trained up to (and including) the TSP-$N$ instances of the curriculum learning. The first training stage of the curriculum (leading to the TSP-100 checkpoint) is the bottleneck in terms of computational needs ($1.5 \cdot 10^6$ samples for 50 epochs), requiring an order of magnitude more training steps than the subsequent trainings. It is noteworthy that the size of instances a checkpoint is trained on does not constrain the size of instances that can be solved: as we will see in the experiments, with a TSP-1000 checkpoint, we predict instances which size ranging from 100 to 7397. 

Later, T2TCO \cite{T2TCO} has introduced a ``gradient based search'' procedure during the inference to improve the quality of the solution. 
Both T2TCO and DIFUSCO generates an initial solution which can be improved by a local search procedure such as 2-opt, or MCTS, or by some sampling procedure. 
\XYZ builds further on the discrete state space version of DIFUSCO and T2TCO. 

Concomitant to our work FT2T \cite{FT2T} a faster version of T2TCO based on consistency model \cite{consistency} has been published. FT2T achieves similar results as ours but requires additional expensive trainings: training both the entire diffusion model and the additional consistency model. Also, our modular approach, and particularly our ``key ingredient 1'' can readily be used to improve solution generation in consistency models such as FT2T without the need of any retraining.

\subsection{Other Constructive Neural Solvers for Combinatorial Optimization}

To avoid the costly generation of the supervised training dataset, other approaches leverage unsupervised learning or reinforcement learning. UTSP \cite{utsp} uses scattering attention GNN \cite{sag} to generate a soft indicator matrix that can be simply transformed into a heatmap. The network parameters are optimized through gradient descent on the tour length augmented with a carefully chosen penalty term. The tour is then refined with an extensive Monte-Carlo tree search. They achieve results similar to DIFUSCO using lower training time.

The majority of neural CO solvers rely on reinforcement learning. They are either autoregressive (incremental construction of the solution), or generating the full solution in one-shot. The former category includes BQ-NCO \cite{bqnco}], POMO \cite{pomo} and SymNCO \cite{symnco} while the latter category includes DIMES \cite{dimes}. DIMES employs a meta-learning framework and a continuous parametrization of the solution space which enables a stable REINFORCE-based training.
In BQ-NCO the state space of the MDP associated with the problem is simplified by bisimulation quotienting. The policy network architecture is based on a transformer \cite{vaswani} or a perceiver \cite{perceiver} architecture.
POMO builds on the attention-based model of \cite{KoolAM} but leverages symmetries by data augmentation and stabilizes REINFORCE training by using multiple initiations. SymNCO also leverages symmetries by introducing regularization in REINFORCE and by learning invariant representation for pre-identified symmetries. SymNCO and POMO do not scale well beyond 200 cities for the TSP. ICAM \cite {ICAM} builds upon POMO, modifying the attention mechanism and the reinforcement learning training scheme to significantly improve performance on larger TSP instances (up to $10^3$ cities). 

Another line of works aims at building solutions by splitting the problem into simpler and smaller sub-problems and merging together the partial solutions. To the best of our knowldged GLOP \cite{glop} is the current SOTA for such methods. 


\section{\XYZ: the Recipe}



Given a random variable $\mathbf{x}_0$ drawn from a certain distribution $p_{complex}$, a variational diffusion model is a latent variable model that progressively adds noise to $\mathbf{x}_0$ following a Markovian forward process, denoted as $q(\mathbf{x_t|x_{t-1}})$. This generates a sequence $\mathbf{x_{1:T}}$ whose conditional joint distribution is recursively defined: $q(\mathbf{x_{1:T}|x_0})=\prod_{t=1}^Tq(\mathbf{x_t|x_{t-1}})$ and such that $q(\mathbf{x_T})\sim p_{simple}$ where $p_{simple}$ is an easy to sample from distribution such as a Gaussian for continuous state diffusion or a multinomial for discrete state diffusion. The reverse process, called the backward process,
$p(\mathbf{x}_{t-1}|\mathbf{x}_t,\mathbf{x}_0)$
is approximated by a Markovian process parameterized by a neural network:
$p_{\theta} (\mathbf{x}_{t-1}|\mathbf{x}_t)$.
The training is performed by searching for $\theta$ that minimizes the variational upper bound of the negative log-likelihood (denoting by $D_{KL}$ the Kullback–Leibler divergence):

\begin{multline}
    L_{vb}=\mathbb{E}_{q(\mathbf{x_0})} [D_{KL}[q(\mathbf{x_T}|\mathbf{x_0}) ||p(\mathbf{x_T)}] \\ + \sum_{t=2}^T\mathbb{E}_{q(\mathbf{x_t | x_0})} [D_{KL}[q(\mathbf{x_{t-1}}|\mathbf{x_t,x_0}) ||p_{\theta}(\mathbf{x_{t-1}}|\mathbf{x_t})]] \\
    - \mathbb{E}_{q(\mathbf{x_1|x_0})} [log~ p_0(\mathbf{x_0}|\mathbf{x_1}) ] ]
    \label{elbo}
\end{multline}

In the case of a discrete diffusion, the equilibrium distribution $p_{simple}$ is induced by the Markov transition matrix $\mathbf{Q}$ associated with the forward transition kernel. Let us denote $x$ the one-hot encoded version of $\mathbf{x}$, we have:

\begin{align*}
  q(x_t|x_{t-1})  =Cat(x_t: p=x_{t-1}\mathbf{Q_t}) \\
  q(x_t|x_0)=Cat(x_t; p=x_0\prod_{i=1}^T\mathbf{Q_i}) = Cat(x_t; p=x_0\mathbf{\bar{Q_t}})\\
  q(x_{t-1}|x_t,x_0)=Cat(x_{t-1}; p =\frac{x_t\mathbf{Q_t}^T \odot x_0\mathbf{\bar{Q}_{t-1}}}{x_0\mathbf{\bar{Q}_t}x_t^T}),
\end{align*}
where $Cat(x;p)$ denotes the categorical distribution with probability vector $p$ and $\odot$ denotes the Hadamard product.

In previous works on combinatorial optimization, $x$ is the adjacency matrix (or the adjacency list if using graph sparsification) of an optimal solution of the TSP instance, which is a $\{0,1\}^{N \times N}$ matrix for a TSP-$N$ (or a $\{0,1\}^{N \times M} $ matrix with $M \in \mathbb{N}, 1\leq M\leq N$ if the graph is sparsified. For notational simplicity, we will now on assume no sparsification. This has no implication on the validity of our presentation).  
The backward transition is parameterized by a neural network which predicts the probability of the target adjacency matrix, which is the heatmap. 
We denote $\Tilde{x_0}$ this predicted adjacency matrix. The conditioning on the instance of the combinatorial optimization problem is implicitly done by the neural network (denoted as $NN_{\theta}$ where $\theta$ is the set of parameters of the network) which is a Graph Neural Network taking as input the graph instance $G$ so that we have $p(\Tilde{x_0} | G)=NN_{\theta}(x_t,t | G)$. 

The heatmap associated with the TSP instance is a $[0;1]^{N \times N}$ matrix. However, the space of $[0;1]^{N \times N}$ matrices is much larger than the space of adjacency matrices associated with Hamiltonian tours. The true solution lies on this much smaller space. Taking into account this fact is the key element 1 of \XYZ that we detail below. Another observation is that a local search procedure is performed on the initial solution generated by the diffusion process (typically 2-opt). Hence any Hamiltonian tour which is transformed into a locally optimal Hamiltonian tour by the local search (2-opt) is equivalent to the latter. This defines a redundant learning objective which we hypothesize will be easier to learn due to this redundancy while not changing the solution space (the set of Hamiltonian tours). This is the key ingredient 2 of \XYZ that we detail below.

\subsection{Key Ingredient 1: Leveraging the Constrained Structure of the Solution Space to Improve Inference-time Solution Generation}

In the TSP, the true $x_0(x_t,t|G)$ lies on a very constrained manifold of optimal Hamiltonian tours. In the original implementation of DIFUSCO, such a constraint is not enforced, the network has to learn it, leading to sub-optimal convergence of the solution to possibly non-Hamiltonian adjacency matrices.
In order to circumvent this problem, we propose to modify the parameterization of the backward process by applying a Hamiltonian tour reconstruction operator $H$ to the predicted $\Tilde{x_0}$ to enforce the Hamiltonian constraint. This tour reconstruction operator is the one used at the end of the backward process in DIFUSCO and T2TCO to reconstruct the tour from the generated heatmap. 
Similarly, to guide the backward process towards an optimal tour, we apply the 2-opt. Let us denote $R_2$ the 2-opt operator. As $H\circ \Tilde{x_0} $ is now a Hamiltonian tour, we can further refine our estimate of $\Tilde{x_0}$ as follows: $\hat{x_0}(x_t,t,G) = R_2 \circ H \circ NN_{\theta}(x_t,t,G)$. 
There is no gradient flow through the $R_2$ operator and the $H$ operator involves deterministic affine transformation of $NN_{\theta}(x_t,t,G)$ which leaves the optimal $\theta$ in equation \ref{elbo} unchanged thus avoiding the need to retrain the entire diffusion model. 

\subsection{Key Ingredient 2: Leveraging Equivalence Class over Target Tours to Define a Redundant Training Objective}

If $x_0$ is the adjacency matrix of an optimal tour then $R_2 x_0=R_2^n x_0 = x_0$ where $R_2^n$ denotes $n$ successive applications of 2-opt: an optimal (locally or globally) tour is a fixed point of 2-opt. Let us further denote by $\mathcal{R}_2$ the following equivalence relation: $x \mathcal{R}_2 y$ if $\exists n \in \mathbb{N}$ such that either $x = R_2^n y $ or  $y = R_2^n x $.
In DIFUSCO the training objective is a Dirac mass located on $x_0$ for a given instance $G$. We propose to replace it by the uniform distribution over the tours from the equivalence class of $x_0$ for the relation $\mathcal{R}_2$.
This procedure increases the cardinality of the set supporting the target distribution which we hypothesize will drive the diffusion process towards a more robust distribution. 
This procedure allows us to leverage the curriculum learning of DIFUSCO where the most resource-consuming steps are the training of the TSP-100 checkpoint. Each of the subsequent training is initialized with the previous checkpoint and it requires approximately an order of magnitude less training steps making these subsequent curriculum learning steps a very good fit for this retraining. 
This requires to sample solutions from the equivalence class of $x_0(G)$ by the equivalence relation $\mathcal{R}_2$ for any instance $I$. This procedure is expensive in terms of computations due to the local minima that can be generated when naively inverting the 2-change several times. To circumvent this problem while keeping the computational time low, we approximate the 2 opt by the 2-change and keep their number small enough ($2$ in all the experiments below). 
This is creating $\mathcal{O}(N^{-1})$ local minima, which is small enough. 

\begin{algorithm}
  \caption{\XYZ inference. $H$ denotes the Hamiltonian reconstruction tour procedure (the one used by DIFUSCO and T2TCO), and $R$ is the 2-opt operator.} 
  \label{alg:IDEQinf}
\begin{algorithmic}
  \REQUIRE The inference schedule (i.e.\@ a sequence of decreasing timesteps from $T$ to $0$: $\mathcal{T}=\{t_T,... t_0=0\}$)
  \REQUIRE $n \in \mathbb{N}$, 
  $0 < \alpha <1$
  \REQUIRE a (trained) neural network $NN_{\theta}$
  \REQUIRE $G$ is the instance to solve. 
  \STATE $x_T \gets$ a one-hot encoded adjacency list randomly sampled from a Bernoulli distribution 
  \FOR{t in $\mathcal{T}\backslash t_0$}
    \STATE $\Tilde{x_0} = R\circ H\circ NN_{\theta}(x_t,t | G)$
    \IF{$t-1 \neq t_0$}
    \STATE Sample $x_{t-1} \sim Cat(x_{t-1}, p=\frac{x_t\mathbf{Q_t}^T \odot \Tilde{x_0 }\mathbf{\bar{Q}_{t-1}}}{\Tilde{x_0 }\mathbf{\bar{Q}_t}x_t^T}) $ 
    \ENDIF
  \ENDFOR
  \FOR{$i$  in $1... n$}
    \STATE sample $x_{\alpha T} \sim Cat(x_{\alpha T} , p = x_0 \mathbf{\bar{Q}_{\alpha T}})$
    \FOR{$t$ in $\{t_{\alpha T},... t_0=0\}\backslash t_0$}
    \STATE $\Tilde{x_0} = R\circ H\circ NN_{\theta}(x_t,t | G)$
    \IF{$t-1 \neq t_0$}
    \STATE Sample $x_{t-1} \sim Cat(x_{t-1}, p=\frac{x_t\mathbf{Q_t}^T \odot \Tilde{x_0 }\mathbf{\bar{Q}_{t-1}}}{\Tilde{x_0 }\mathbf{\bar{Q}_t}x_t^T}) $ 
    \ENDIF
    \ENDFOR
  \ENDFOR
  \RETURN $\Tilde{x_0}$
\end{algorithmic}
\end{algorithm}

\subsection{\XYZ: Training and Inference}

Building on these two ingredients, let us specify the training and the inference phases of \XYZ:

\begin{itemize}
\item \textbf{Training phase:} 
  \begin{itemize}
  \item to save computation time, we initialize $NN_{\theta}$ with the DIFUSCO original TSP-100 checkpoint.
  \item Then, we train an \XYZ TSP-500 checkpoint with the training objective modified as follows: instead of the optimal tour, we use a tour obtained by applying 2 consecutive randomly sampled 2-change on the optimal tour (key ingredient 2). This produces the \XYZ TSP-500 checkpoint. 
  \item Likewise for the TSP-1000 checkpoint: we initialize the network with the \XYZ TSP-500 checkpoint and train it with the updated objective (key ingredient 2). This produces the \XYZ TSP-1000 checkpoint.
  \end{itemize}
\item \textbf{Inference phase:} we follow the inference steps of T2TCO: the diffusion goes from $t=T$ to $t=0$ and is then  followed by $n$ ($n \in \mathbb{N}$ is a hyper-parameter) partial forward/backward diffusion steps from $t=0$ to $t=\alpha T$ and vice versa (where $0< \alpha <1$ is a hyper-parameter) but we replace the standard diffusion steps by the updated diffusion steps detailed in algorithm \ref{alg:IDEQinf}.
\end{itemize}


\section{\XYZ: Experimental Study}

In computational CO, experimental results are key. Henceforth, in this section, we detail our experimental results. The most important result is that \XYZ reaches SOTA performance on the TSPlib benchmark, achieving  significantly better results than the previous SOTA DIFUSCO and T2TCO. 
Overall, \XYZ exhibits state-of-the-art performance for a neural network approach on the TSP.

\begin{table*}[t]
  \centering
    \caption{Comparison of averaged optimality gaps (OG) and running times on large scale 2D uniformly distributed Eulidean TSP instances. UL = Unsupervised learning, RL = Reinforcement learning, SL = supervised learning, G = Greedy decoding, aug.\@ = data augmentation, S = sampling, BS = beam search. For both sampling and beam search, we use the notation 'xS' to indicate a sampling size of S or a beam search of size S. The results marked with an asterisk (*) are obtained by running ourselves the code, others originate from the publications. 'ref.' indicates the reference method used to calculate the optimality gap. Information about the variance of optimality gaps is available in figure \ref{fig:var} and table \ref{table:stddev}.}
  \begin{tabular}{l|c||c|c||c|c|}
    \multicolumn{2}{c||}{Algorithm} & \multicolumn{2}{c||}{TSP 500} & \multicolumn{2}{c|}{TSP 1000}  \\
    Name & Type & OG. & running time & OG. & running time   \\
    \hline
    Concorde (*) &Exact &  0.0~\% (ref.) & 2.6~mn &  0.0~\% (ref.) & 1.2~h \\
    LKH3 (*) &Heuristic &  0.0~\% & 0.2~mn &  0.2~\% & 0.4~mn \\
    2-opt (*) &Heuristic &  13~\% & 0.6~s   &  13~\% & 13~s \\    
    \hline
    POMO x8 & RL + S  & 22~\% & 1.1~mn & 41~\% & 8.5~mn \\
    GLOP  & RL & 2.2~\% & 1.6~mn & 3.1~\% & 3.3~mn \\
    ICAM & RL & 1.6~\% & 1~s & 2.9~\% & 2~s \\
    ICAM aug. x8 & RL & 0.77~\% & 0.6~ mn & 1.6~\% & 3.8~mn\\
    BQ-NCO & RL & 1.2~\% & 0.9~mn & 2.3~\% & 2~mn \\
    BQ-NCO x16 & RL +BS  & 0.55~\% & 15~mn & 1.4~\% & 38~mn \\
    UTSP & UL + MCTS &0.8~\%  & 2.7mn  & 1.2~\% & 6.0~ mn \\
    DIFUSCO (*) & SL+G+2-opt & 1.6~\% &0.75~mn & 2.0~\% &  0.8~mn \\
    DIFUSCO (*) x4 & SL+S+2-opt& 0.93~\% & 2.6~mn & 1.5~\% & 3.3~mn \\ 
    T2TCO (*) & SL+G+2-opt & 0.85~\% & 0.9~mn & 1.5~\% & 3.0~mn \\
    T2TCO (*) x4 & SL+S+ -opt& 0.67~\% & 1.5~mn & 1.2~\% & 4.9~mn \\ 
    \hline
    \XYZ (*) & SL+G+2-opt &\textbf{0.41~\%} & 1.3~mn & \textbf{0.63~\%} & 2.3~mn \\
    \XYZ (*) x4 & SL+S+2-opt& \textbf{0.29~\%} & 3.7~mn & \textbf{0.49~\%} & 6.1~mn \\
  \end{tabular}
  \label{main_res}
\end{table*}

\subsection{Experimental Setup and Reproducibility}

The code to reproduce the experiments in the paper, including dataset generation can be found on the accompanying website (\url{https://github.com/mkbas/IDEQ}). Hyperparameter values can be found on this website.

We did not rerun all the algorithms, only DIFUSCO and T2TCO. Indeed, we tried to be mindful of resource use and avoid duplicating already published experiments. However, due to their closeness with IDEQ that makes performance and running time comparisons critical, we re-reran DIFUSCO and T2TCO on the same hardware as IDEQ. Running times and optimality gaps of POMO, LKH3, ICAM and GLOP are taken from \cite{ICAM} where they were run on the same hardware, while for BQ-NCO and UTSP they are taken from their respective publications \cite{bqnco,utsp}. 
In all these studies, the test sets comprised 128 random instances for TSP-500 and TSP-1000. However, our experiments showed that the variance of the results was high so we increased the size of our test set to 2048 random instances. We then scaled down running times to 128 ($/16$) to ensure a fair comparison (the running time is linear in the size of the test set in this setting). 
Following their original publications, the inference for DIFUSCO, T2TCO and consequently \XYZ is done with pytorch-based code with a batch size of 1. When performed, sampling is done sequentially. 

The experiments based on neural networks were run on a cluster of 4 Nvidia A100 40GB GPUs with AMD EPYC 7513 CPU. Concorde and LKH3 are highly optimized C code. We ran these on a 2x Intel Xeon Platinium 8358 32-core CPU server. The 2-opt algorithm is initiated on a random tour and ran on the same hardware as Concorde.

\subsection{Experimental Methodology}

First, we compare \XYZ to other approaches on 2D Euclidean instances made of $N$ cities drawn uniformly at random, provided by the authors of DIFUSCO. 
However, real TSP instances are not random in this way; instances are structured because they correspond to real-world problems. It is well-known in CO that the performance on uniformly random instances does not say much about the performance on structured instances. With this in mind, we ran \XYZ and other diffusion-based approaches on the TSPlib \cite{tsplib}, a highly renown benchmark in the TSP community. We used the TSPlib instances from $10^2$ to $10^4$ cities. As DIFUSCO does not deal with non Euclidean instances, we converted the few geographic instances of the TSPlib to Euclidean and solved them with Concorde for instances up to $10^3$ cities, and larger instances with LKH3.
Finally, to better understand how \XYZ works, we performed ablation studies.
Results are reported in the next 2 sections.

\subsection{Experimental Results}

Table \ref{main_res} reports on the first experiments in which we compare the performance of the different approaches on random instances.
\XYZ clearly exhibits smaller optimality gaps as well as a better scaling behavior when going from 500 cities instances to 1000 cities instances.

On the TSPlib, we used the TSP-1000 checkpoint to solve all instances. 
Table \ref{tsplib_res} shows both the consistency of \XYZ performance and its ability to generalize beyond the training distribution. 

\begin{table}[t]
  \caption{Comparison of optimality gaps (OG) and running times on TSPlib instances ranging from 100 to $10^4$ cities.}
  \centering
  \begin{tabular}{l||c|c|}
    & \multicolumn{2}{c|}{TSPlib 100-$10^4$}    \\
    Algorithm & OG. & running time  \\
    \hline
    DIFUSCO   & 2.9~\% & 1.5~mn \\
    T2TCO & 1.9~\% & 4.5~mn \\
    \XYZ &  1.1~\% &  4.0~mn\\
    \hline
    DIFUSCO x4 & 1.9~\% & 5.3~mn  \\ 
    T2TCO x4 &  1.7~\% &  8.1~mn \\
    \XYZ x4 & 0.89~\% & 9.8~mn  \\
  \end{tabular}

  \label{tsplib_res}
\end{table}    

\begin{figure}[!ht]
    \centering
    \includegraphics[width=0.6\linewidth, trim={0.5cm 1.2cm 0.5cm 1.2cm},clip]{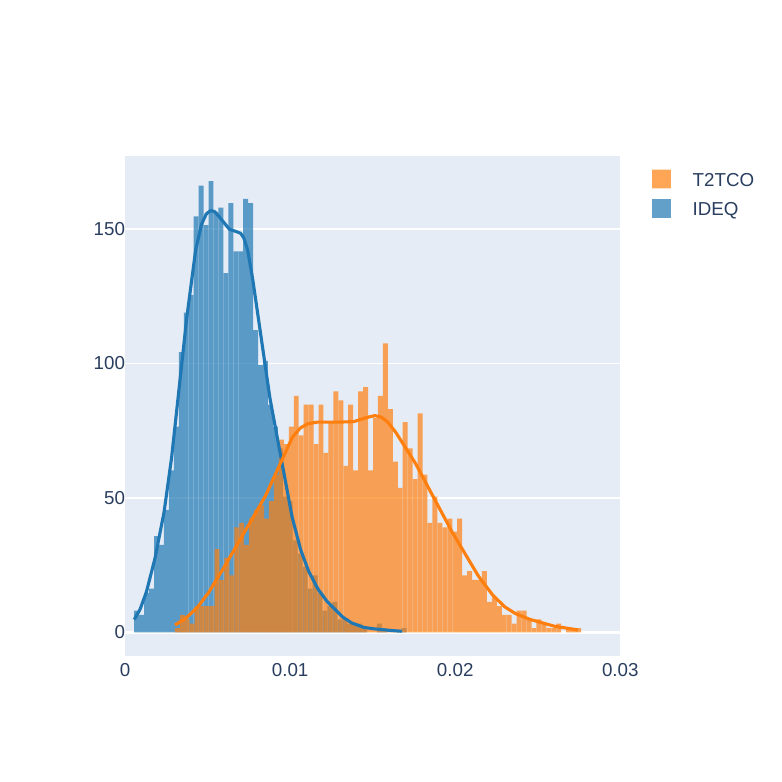}
    \caption{Distribution of optimality gaps (x-axis) measured on 32 repetitions of the first 64 TSP-1000 instances from the test set. \XYZ in blue and T2TCO in orange. Bars represent the empirical data distribution and the line is the Gaussian kernel density estimate.}
    \label{fig:var}
\end{figure}

Finally, we explored the variability of the results, exploring both the sampling-driven variability for a given instance and the optimality gap on different instances. Indeed, for a given instance, the denoising process can start from various noisy graphs with no guarantee to converge to the same solution. We therefore looked at 32 random initial tours for 64 different TSP-1000 instances and looked at the distribution of the predicted solutions and associated optimality gaps. The results are shown in figure \ref{fig:var}.
The standard deviations of the optimality gap obtained on 2048 instances of the TSP-1000 are shown in table \ref{table:stddev}. 
\XYZ exhibits a smaller variability for the length of the inferred tour, both within and between instances, which is consistent with the reduced search space and the more robust objective retraining.

\begin{table}
  \caption{Standard deviation of the optimality gap of the tours generated by 4 different approaches on TSP 1000 instances. Compared to T2TCO, \XYZ more than halves the standard deviation. OG stands for “optimality gap” it is reminded here to ease the interpretation of the standard deviation.}
  \centering
  \begin{tabular}{c|c|c}
    Algorithm & Mean std.\@ dev.\@  &OG.\\ \hline
    T2TCO & $0.13$ & 1.4~\%\\
    \XYZ without key ingredient 1  & $0.082$ & 1.0~\%\\
    \XYZ without key ingredient 2 & $0.075$ & 0.93~\%\\
    \XYZ  & $0.057$ & 0.63~\%\\
  \end{tabular}

  \label{table:stddev}
\end{table}

\subsection{Ablation Studies}

We conducted ablation studies on the two components of \XYZ: the updated estimator of $\hat{x_0}(x_t,t,I)$ (key ingredient 1) and the re-training of the later stages of DIFUSCO curriculum learning (key ingredient 2). The ablation of both brings back to T2TCO but without the gradient search.
Removal of the late stage re-training was explored by using the original DIFUSCO checkpoints while keeping the Hamiltonian tour reconstruction operator and the 2-opt at inference time. This is labeled as ``\XYZ without key ingredient 2'' in tables \ref{table:stddev} and \ref{ablation_table}. 
We also studied the effect of the removal of Hamiltonian tour reconstruction operator and the 2-opt during the diffusion process but keeping the \XYZ retrained checkpoint. This is labeled as ``\XYZ without key ingredient 1'' in tables \ref{table:stddev} and \ref{ablation_table}. 
Finally we looked at performance of DIFUSCO with the newly trained \XYZ checkpoint. 

Table \ref{ablation_table} shows the results of these ablation studies. These results are consistent with the two effects adding up independently as we could anticipate based on the methodology.  

\begin{table*}[t]
  \caption{Results of the ablation study of \XYZ. OG stands for ``optimality gap''.}
  \centering
  \begin{tabular}{l||c|c||c|c|}
    &  \multicolumn{2}{|c||}{TSP 500} & \multicolumn{2}{|c|}{TSP 1000}  \\
    Algorithm &  OG. & running time & OG. & running time   \\
    \hline
    \XYZ  & \textbf{0.41~\%} & 3.7~mn & \textbf{0.63~\%} & 5.1~mn \\
    \XYZ without key ingredient 1 & 0.6~\% & 0.4~mn & 1.0~\% & 1.6~mn \\
    \XYZ without key ingredient 2 & 0.65~\% & 1.3~mn & 0.93~\% & 6.2~mn \\
    \XYZ without key ingredients 1 and 2 & 0.89~\% & 0.4~mn & 1.4~\% & 1.6~mn \\
    T2TCO & 0.83~\% & 0.9mn & 1.5~\% & 4mn \\
    DIFUSCO with \XYZ checkpoint & 1.3~\% & 0.7~mn & 1.6~\% & 0.8~mn \\
  \end{tabular}

  \label{ablation_table}
\end{table*}

This also shows that either one of these 2 components alone is outperforming existing SOTA. 


\section{Conclusion and Future Work}

Building on DIFUSCO and T2TCO, we have introduced \XYZ which leads to a significant improvement of the state-of-the-art neural diffusion-based solver for the Traveling Salesman Problem. The optimality gap is significantly reduced on instances of up to thousands of cities. It is noticeable that the optimality gap obtained with \XYZ does not degrade as much as the methods it gets inspired of. It is also noticeable that the variability of the optimality gap of the tours found by \XYZ is smaller than the one of previous neural methods. These modifications rely on an improvement of the inference process, as well as a fine-tuning of the existing checkpoints. As such \XYZ does not require the expensive retraining of a diffusion model from scratch. 
\XYZ relies on a strong effort to understand previous approaches which led us to propose two original key ingredients. 
These two ingredients can be used independently or jointly to improve a wide variety of diffusion models including consistency models with no to minimal retraining needs.  

Our objective was to explore the effect of taking into account the constrained structure of the solution space and the various steps of the pipeline (diffusion model + local search +/- sampling) to improve the quality of the solutions. Our results show the dramatic improvement obtained by taking these into account.
This research is applicable beyond the field of combinatorial optimization. 

\XYZ may be applied to other combinatorial optimization problems for which a simple invertible transformation exists (like the 2-change for the TSP). This application to other combinatorial optimization problems is one line for future work.

In \XYZ, to constrain the diffusion to predict a solution on the space of locally optimal Hamiltonian tours, we used projections. This may not be the most effective way and this is another line of future research.  


\section*{Acknowledgments}
Micka\"el Basson is also a full-time employee of Lilly France detached to work on this project. 
We acknowledge the Région Hauts-de-France CPER project CornelIA. Experiments presented in this paper were carried out using
the Grid’5000 testbed (https://www.grid5000.fr). 
Both authors acknowledge the Scool research group for an outstanding working environment.
\bibliographystyle{splncs04}
\bibliography{bibliography}

\end{document}